\documentclass{article}

\PassOptionsToPackage{numbers, compress}{natbib}

\usepackage[preprint]{neurips_2023}

\usepackage[utf8]{inputenc} 
\usepackage[T1]{fontenc}    
\usepackage{hyperref}       
\usepackage{url}            
\usepackage{booktabs}
\usepackage{amsfonts}
\usepackage{nicefrac}
\usepackage{microtype}
\usepackage{xcolor}
\usepackage{wrapfig}
\usepackage{cutwin}
\usepackage{lipsum}
\usepackage{enumitem}
\usepackage{verbatim}
\usepackage{xspace}
\usepackage{diagbox}
\usepackage{array}
\usepackage{cite}

\usepackage{multirow}
\usepackage{diagbox}
\usepackage{url}
\usepackage{comment}
\usepackage{booktabs}
\usepackage{amsmath}
\usepackage{scalerel}
\usepackage[figuresright]{rotating}
\usepackage{threeparttable}
\usepackage{algorithm}
\usepackage{algorithmic}
\usepackage{siunitx}
\usepackage{stackengine}

\usepackage{tcolorbox}
\definecolor{grey}{RGB}{128,128,128}

\usepackage{mathtools}
\usepackage{multirow}
\usepackage{multicol}
\usepackage{booktabs}
\usepackage{subfigure}

\usepackage{amsmath,amsfonts,bm}

\def\eqref#1{equation~\ref{#1}}

\def\1{\bm{1}}

\DeclareMathAlphabet{\mathsfit}{\encodingdefault}{\sfdefault}{m}{sl}
\SetMathAlphabet{\mathsfit}{bold}{\encodingdefault}{\sfdefault}{bx}{n}

\title{Multi-AGV Path Planning Method via Reinforcement Learning and Particle Filters}

\author{%
Shuo Shao  \\
Wuhan University of Technology \\
\texttt{shaoshuo@whut.edu.cn}
}

\begin{document}

\maketitle

\begin{abstract}
  Thanks to its robust learning and search stabilities,the reinforcement learning (RL) algorithm has garnered increasingly significant attention and been exten-sively applied in Automated Guided Vehicle (AGV) path planning. However, RL-based planning algorithms have been discovered to suffer from the substantial variance of neural networks caused by environmental instability and significant fluctua-tions in system structure. These challenges manifest in slow convergence speed and low learning efficiency. To tackle this issue, this paper presents a novel multi-AGV path planning method named Particle Filters - Double Deep Q-Network (PF-DDQN)via leveraging Particle Filters (PF) and RL algorithm. Firstly, the proposed method leverages the imprecise weight values of the network as state values to formulate thestate space equation.
 Subsequently, the DDQN model is optimized to acquire the optimal true weight values through the iterative fusion process of neural networksand PF in order to enhance the optimization efficiency of the proposedmethod. Lastly, the performance of the proposed method is validated by different numerical simulations. The simulation results demonstrate that the proposed methoddominates the traditional DDQN algorithm in terms of path planning superiority andtraining time indicator by 92.62\% and 76.88\%, respectively. Therefore, the proposedmethod could be considered as a vital alternative in the field of multi-AGV path planning.

\end{abstract}

\section{Introduction}\label{sec1}

Due to its significant role and wide application prospecting in both military and civilian domains, the Autonomous Guided Vehicle (AGV) has gained considerable attention in the field of robotics \citep{b1}. As one of the key factors to achieve high-level intelligence of AGV, Path planning, which is essential for achieving AGV intelligence, has also attracted extensive research interest \citep{b2}. However, path planning of AGV has been discovered and proven to be a NP-hard issue, leading to find an optimal path within a given computation time to be greatly difficult or even impossible. Thus, developing high performance path planning method is always heavily demanded in the filed of AGV path planning.

Generally, AGV path planning algorithms can be categorized into two types, that is,the traditional algorithms and population-based intelligent optimization algorithms. Despite being capable of offering promising performance regarding to the computation time, traditional path planning methods, such as graphical methods\citep{b3}, artificial potential field methods \citep{b4},and dynamic window approaches, still suffer from deficiencies in planning stability and adaptability\citep{b5}. On the other hand, artificial intelligence optimization algorithms, including genetic algorithms \citep{b6}, neural networks \citep{b7}, and ant colony algorithms \citep{b8}, can achieve formidable planning stability in complex environments due to their population-based characteristics \citep{b9}.

As one of the most prominent population-based intelligent optimization algorithms,Reinforcement Learning (RL) exhibits powerful machine learning capabilities and enables the intelligent agent to learn specific behavioral norms through trial and error \citep{b10}. Moreover, this algorithm possesses the characteristics of reward-based feedback and is independent on training data, leading it to be naturally adaptable in the field of path planning \citep{b11}.

RL is an influential machine learning approach that enables intelligent agents to acquire knowledge on how to act within an environment to accomplish predefined objectives through reward feedback. It does so without any prior understanding of the environment or training data \citep{b12}. This characteristic makes it especially well-suited for tasks related to local path mapping. Double Deep Q-Network (DDQN) represents a renowned RL algorithm that has gained extensive utilization in the field of path planning.

However, AGV planning remains challenging for current Deep Reinforcement Learning (DRL) methods, partly due to their need to the following reasons:(i)partially observable environments and (ii) reason through complex observations, such as avoiding collisions between AGVs. While most existing RL path planning algorithms have shown feasibility in single AGV path planning problems, they fail to achieve multi-AGV path planning in more complex environments. This could be probably interpreted by the fact that the neural networks have high variance, leading to slow convergence of the algorithm and reducing its ability to handle cooperative planning among multiple agents in complex environments \citep{b13}.

To handle the issue noted above, some studies have proposed many different improved RL-based planning methods by increasing the complexity of the grid state space and incorporating perception state space \citep{b14}. However, this type of method still relies on training results in an ideal noise-free environment, resulting in the inability to address the issue of weight inaccuracy \citep{b15}.
While feasible, overcoming these constraints incurs additional computational expenses when training deep neural networks.Additionally, some research has aimed to improve DRL algorithms by introducing Kalman Filters (KF) \citep{b16}. However, traditional KF require linearization of the model through Taylor series expansion, which compromises the accuracy of the model to some extent, as they are only effective for linear systems.

This study commits its interest on multi-AGV path planning in an unfamiliar
environment. In typical scenarios, intricate tasks involving mobile robots, such as mapping, item delivery, or surveillance, necessitate path planning for multiple targets. Our conjecture is that the conventional DDQN approach fails to address the challenge of path planning for multiple targets due to the convergence of Q-values solely based on the spatial coordinates on the map. Empirical findings illustrate that while DDQN can successfully learn to plan a path for the initial target, it struggles to efficiently plan for subsequent targets due to the environment's complexity and the incapability to learn and regenerate the value network for the remaining targets. Despite researchers' efforts to modify DDQN and reduce convergence time, there is currently no existing research on employing PF integration to tackle the issue of multi-AGV path planning within this environment. The subsequent section will provide a more comprehensive explanation of the DDQN algorithm.

To address the aforementioned issues, this paper proposes a PF-DDQN-based method for multi-AGV path planning. Firstly, in the proposed method, the training network with environmental noise and the target network with inaccurate weights are treated as state and observation variables, respectively, to construct the system's state equation and observation equation. Then, through the fusion iteration of neural networks and PF, the weights of the neural network are continuously updated to improve the convergence speed of the proposed algorithm. Finally, the effectiveness and superiority of the proposed method are validated through numerical simulations under different operating conditions.

The rest of this paper is as follows: Section 2 introduces related work. Section 3 presents the theoretical background of methods. Section 4 compares the experimental results of the proposed algorithm with DDQN and EKF-DDQN. Finally, Section 5 provides a summary of this paper.

\section{Related Work}

Genetic Algorithms (GAs) and Ant Colony Optimization (ACO) are heuristic methods that show potential in solving optimization problems \citep{b17}. GAs generate a collection of potential solutions through Darwinian evolution principles and assess them using a fitness function. The fittest individuals are selected for reproduction through crossover operations, while population diversity is maintained through mutation operations \citep{b18}. Conversely, ACO draws inspiration from the behavior of ants and utilizes pheromone trails to improve the efficiency of the shortest paths between initial and final points \citep{b19}.

In recent years, researchers have proposed various optimization techniques for path planning of mobile robots in grid environments. Li et al. \citep{b20} introduced an enhanced algorithm called Improved Ant Colony Optimization-Improved Artificial Bee Colony (IACO-IABC), which effectively improves the performance of mobile robot path planning by combining the strengths of both algorithms and introducing new heuristics and search mechanisms. Li et al. \citep{b21} proposed an advanced Ant Colony Optimization (MACO) algorithm that addresses issues such as local optima and slow convergence in trajectory planning for Unmanned Aerial Vehicles (UAVs) by incorporating the metropolis criterion and designing three trajectory correction schemes. Additionally, they employed an Inscribe Circle (IC) smoothing method to enhance efficiency and safety in trajectory planning.

While GAs and ACO have emerged as solutions for path planning problems, their effectiveness in real-time path planning is limited. These population-based algorithms are computationally demanding, particularly for large search spaces, making them unsuitable for real-time applications that require prompt decision-making. Furthermore, the population-based approaches employed by these algorithms can yield suboptimal solutions and exhibit slow convergence, especially in complex environments \citep{b22}.

The Double Deep Q-Network (DDQN) algorithm has gained significant traction in the field of reinforcement learning for acquiring optimal action sequences based on a predefined policy. This algorithm is renowned for its simplicity, speed, and effectiveness in addressing challenging problems in complex and unknown environments. Numerous researchers have applied the DDQN algorithm to path planning by discretizing real-world scenarios into a state space and subsequently simulating them. These methods differ in terms of task-specific behavior design and the representation of the state space.

One of the earliest approaches was the Deep Q-Network (DQN) algorithm proposed by Mnih et al. \citep{b23}, which combines deep neural networks with Q-learning to address reinforcement learning problems in high-dimensional and continuous state spaces. The DQN algorithm resolves the instability issue in Q-learning by employing an experience replay buffer and a target network.

To address the overestimation problem in the DQN algorithm, Hasselt et al. \citep{b24} introduced the use of two Q-networks: one for policy action selection and the other for action value estimation. They utilized a target value network to alleviate the overestimation problem and improve training stability. They also introduced distributed reinforcement learning, modeling the value of actions as probability distributions rather than single values. The use of quantile regression to estimate the distribution of action values enhances learning efficiency and stability \citep{b25}.

Igl et al. \citep{b26} adopted the Variational Sequential Monte Carlo method, while Naesseth et al. \citep{b27} applied PF for belief tracking in reinforcement learning. The latter method enhances belief tracking capability, but experiments reveal that generative models lack robustness in complex observation spaces with high-dimensional uncorrelated observations. To address this issue, more powerful generative models such as DRAW \citep{b28} can be considered to improve observation generation models. However, evaluating complex generative models for each particle significantly increases computational costs and optimization difficulty.

Recent attention has focused on embedding algorithms into neural networks for end-to-end discriminative training. This idea has been applied to differentiable histogram filters, such as Jonschkowski \& Brock \citep{b29}, and KF \citep{b30}. Ma et al. \citep{b31} integrated PF with standard RNNs (e.g., LSTM) and introduced PF-RNN for sequence prediction.

Muruganantham et al. \citep{b32} proposed a dynamic Multi-Objective Evolutionary Algorithm (MOEA) based on Extended Kalman Filters (EKF) prediction. These predictions guide the search for changing optima, accelerating convergence. A scoring scheme was designed to combine EKF prediction with random reinitialization methods to enhance dynamic optimization performance.

Gao et al. \citep{b33} presented an adaptive KF navigation algorithm, RL-AKF, which employs reinforcement learning methods to adaptively estimate the process noise covariance matrix. Extensive experimental results demonstrate that although this algorithm accurately estimates the process noise covariance matrix and improves algorithm robustness, it is time-consuming and faces convergence challenges.

In this work, we depart from directly combining PF and RL algorithms. Instead, building upon previous work, we treat the weightsof the neural network as state variables in the PF algorithm. We leverage the highly adaptive and nonlinear iterative characteristics of PF to update the network weights, reducing the variance of the target network in the RL algorithm. This approach enhances the accuracy of action guidance for trajectories and improves the convergence speed of the reinforcement learning algorithm.

\section{Methods}

\subsection{Particle Filtering}

PF is a recursive Bayesian filtering algorithm utilized for state estimation within a system. Its core principle involves approximating the probability distribution $p(x_k|z_{1:k})$ by employing a set of random samples, known as particles, which represent the states \citep{b18}. Here, let $x_{0:k} = {x_0, x_1, \ldots, x_k}$ and $z_{0:k} = {z_0, z_1, \ldots, z_k}$ represent the target states and observed values at time $k$. The particle weight update function depends on the target states and observation data $x_{i,k} \sim q(x_{i,k} \mid x_{i,k-1}, z_k)$, which can be expressed as follows:

\begin{equation}
  w_k^i = \frac{p(z_k|x_k^i)p(x_k^i|x_{k-1}^i)}{q(x_k^i|x_{k-1}^i,z_k)} \cdot w_{k-1}^i
\end{equation}

Consequently, a collection of particles with weight values ${x_k^i,w_k^i}_{i=1}^{N_s}$ approximates the posterior probability density, where the importance sampling function is defined as follows:

\begin{equation}
  q\left(\frac{x_k^i}{x_{k-1}^iz_k}\right) = p\left(\frac{x_k^i}{x_{k-1}^i}\right)
\end{equation}

The PF algorithm follows the subsequent steps:

Step 1: Sample $x_{i,k}\sim q(x_{i,k}|x_{i,k-1},z_k)$ for $i=1,\ldots,N_s$.

Step 2: Calculate the importance weight $\tilde{w}_k^i$, normalize it, and update particle weights.

Step 3: Compare the effective sample size $\tilde{N}{\mathrm{eff}}$ with a predefined threshold $N{\mathrm{th}}$. If $\tilde{N}{\mathrm{eff}} < N{\mathrm{th}}$, perform resampling of $({x_{k}^i,w_k^i})_{i=1}^{N_s}$.

Unlike the conventional KF, PF is not reliant on linearization or Gaussian assumptions, making it more suitable for the iterative updating of neural networks \citep{b19}.

\subsection{Extended Kalman Filtering}

The Extended Kalman Filter (EKF) is an extension of the KF that addresses the estimation of system states in the presence of nonlinear system dynamics and measurement models. It approximates the nonlinear models using linearization techniques and applies the EKF update equations to iteratively estimate the state of the system.

Similar to the KF, the EKF maintains an estimate of the system state, denoted as $\hat{x}_k$, and an error covariance matrix, denoted as $P_k$, which represents the uncertainty associated with the estimate. The EKF incorporates new measurements to update the state estimate and the error covariance matrix.

The EKF consists of two main steps: the prediction step and the update step.

\textbf{Prediction Step:}
In the prediction step, the EKF predicts the current state based on the previous estimate and the nonlinear system dynamics. It updates the estimate of the state and the error covariance matrix as follows:

\begin{align}
  \hat{x}_k^- & = f(\hat{x}_{k-1}, u_{k-1}) \\
  P_k^-       & = F_kP_{k-1}F_k^T + Q_k
\end{align}
where $\hat{x}_k^-$ and $P_k^-$ are the predicted state estimate and the predicted error covariance matrix, respectively. The function $f(\cdot)$ represents the nonlinear system dynamics, and $u_{k-1}$ is the control input at time step $k-1$. The matrix $F_k$ is the Jacobian matrix of the system dynamics function evaluated at the predicted state, and $Q_k$ is the process noise covariance matrix that represents the uncertainty in the system dynamics.

\textbf{Update Step:}
In the update step, the EKF incorporates the new measurement to improve the estimate of the state. It computes the Kalman gain, which determines the weight given to the measurement and the predicted state estimate, and updates the state estimate and the error covariance matrix as follows:

\begin{align}
  K_k       & = P_k^-H_k^T(H_kP_k^-H_k^T + R_k)^{-1}    \\
  \hat{x}_k & = \hat{x}_k^- + K_k(z_k - h(\hat{x}_k^-)) \\
  P_k       & = (I - K_kH_k)P_k^-
\end{align}
where $K_k$ is the Kalman gain, $H_k$ is the Jacobian matrix of the measurement function evaluated at the predicted state, $R_k$ is the measurement noise covariance matrix, and $z_k$ is the measurement obtained at time step $k$. The function $h(\cdot)$ represents the measurement model that relates the true state to the measurements.

The EKF linearizes the nonlinear system dynamics and measurement models using the Jacobian matrices. By approximating them as linear, the EKF can apply the KF update equations to estimate the state. However, the linearization introduces errors, and the accuracy of the EKF depends on the quality of the linearization. In cases where the nonlinearities are significant, more advanced techniques like the Unscented KF or PF may be more appropriate for state estimation.

\subsection{Double Deep Q-network Algorithm}

The DDQN algorithm is an enhanced iteration of the Deep Q-Network (DQN) algorithm \citep{b21}. In contrast to conventional methods with similar objectives, DDQN mitigates the adverse effects of overestimation by employing the current estimation network to approximate the maximum Q-value of the subsequent state, rather than relying solely on the target network \citep{b23}. Within the DDQN algorithm, a target network architecture is established to minimize the loss function, which can be mathematically expressed as follows:

\begin{equation}
  L(\theta) = \frac{1}{2}\sum_i \left(Q(s_i,a_i;\theta) - y_i\right)^2
\end{equation}
where $y_i$ represents the target Q-value, and $Q(s_i,a_i;\theta)$ represents the predicted Q-value. The network is trained using stochastic gradient descent to update the network weights $\theta$ at each time-step $i$, resulting in improved estimates of the Q-values. The update expression of the Q-value as follows:

\begin{equation}
  Q(s_i,a_i;\theta) = r_t + \gamma \max Q(s_{t+1},a_{t+1}|\tilde{\theta})
\end{equation}
where $r_t$ represents the reward value obtained from the corresponding action at time $t$, while $\gamma$ denotes the discount factor. The derivative of the loss function with respect to the network parameters $\theta$, in terms of $\gamma$, is expressed as follows:

\begin{equation}
  \nabla_\theta L(\theta) = \sum_i \left(Q\left(s_i,a_i,\theta\right) - y_i\right)\nabla_\theta Q\left(s_i,a_i;\theta\right)
\end{equation}

At this juncture, the update rule for the weights is defined as illustrated in following Equation:

\begin{equation}
  \theta_{t+1} = \theta_t + \alpha^(Z_t-Q(s_t,a_t;\theta)\frac{dQ(s_t,a_t;\theta)}{d\theta_t})
\end{equation}
where $\theta_t$ represents the network weights at the current moment, while $\theta_{t+1}$ denotes the updated network weights. $\alpha$ signifies the learning rate, which governs the step size of weight updates. $Z_t$ represents the value of the target network's value function for the current state and action \citep{b25}.

The DDQN algorithm improves convergence speed and addresses overestimation issues in traditional DQN algorithms. However, it still faces challenges in effectively addressing weight inaccuracy within the model, which limits its superiority. To overcome these limitations, this study introduces a novel algorithm called PF-DDQN, which combines PF and DDQN.

\subsection{Simulation Environment Modeling}

\begin{figure}[t!]
  \centering
  \includegraphics[scale = 0.3]{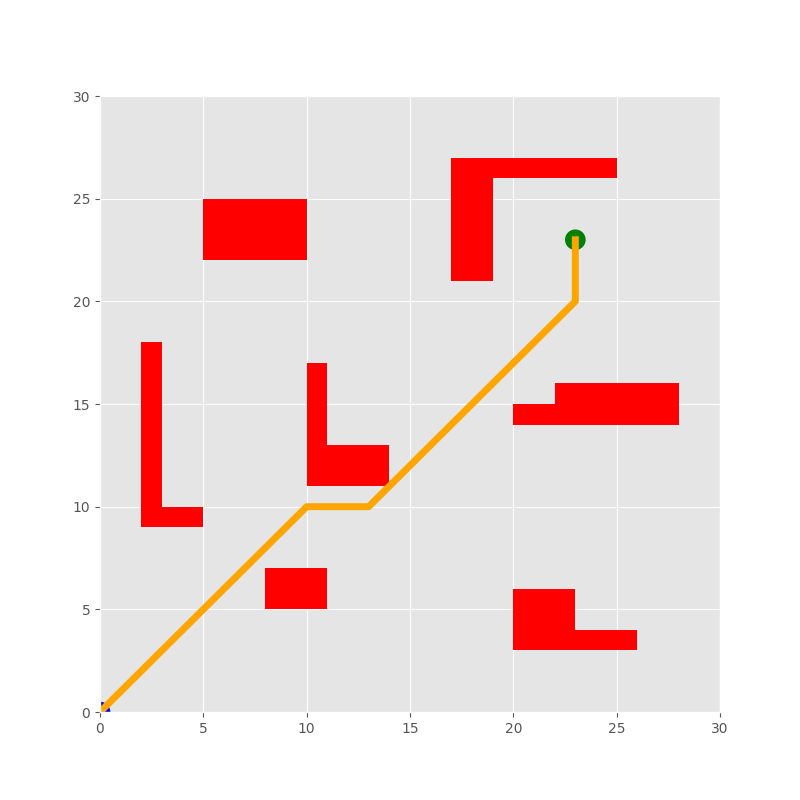}
  \caption{\textbf{The 2D map of AGV path planning}}
  \label{The 2D map of AGV path planning}
\end{figure}

The environment is a 2D map as shown in Figure \ref{The 2D map of AGV path planning}, represented as a 30×30-pixel image, where each pixel corresponds to a 2-meter size in the real world. The training utilizes an entirely unfamiliar map, which is a 30×30-pixel image. In this image, the red and white regions correspond to obstacles and free space, respectively. The trajectory of the AGV and the target are depicted by different colors, with the former represented by colors other than green. Since the mobile robot lacks knowledge of the reference map, it is imperative for it to generate its own map.

The goal of AGV spatial trajectory planning is to achieve collision-free detection for the AGV. Therefore, during spatial trajectory planning, optimization of the trajectory is required based on two aspects: the relationship between the AGV and obstacle positions and the relative position between the AGV and obstacles.

\begin{figure}[t!]
  \centering
  \includegraphics[scale = 1.0]{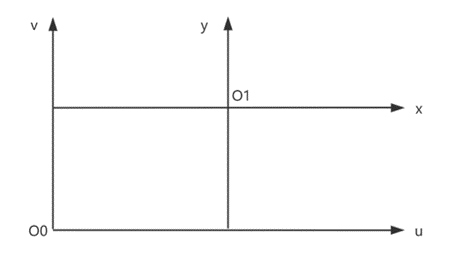}
  \caption{\textbf{Graph environment image coordinate system}}
  \label{Graph environment image coordinate system}
\end{figure}

Real AGVs rely on vision to determine their position. In this work, the three-dimensional position parameters of the target are obtained through trigonometry and disparity. Given the relative positioning between two cameras and the internal parameters of the cameras, the spatial coordinates or dimensions of an object can be obtained by knowing the disparity of its features. The coordinate system of the environmental image is shown in Figure \ref{Graph environment image coordinate system}.

As depicted in Figure \ref{Graph environment image coordinate system}, a u-v-o Cartesian coordinate system is defined in the obtained image, where the pixel coordinates correspond to the rows and columns in the sequence. Therefore, (u, v) represents the pixel coordinate system of the AGV.

Since (u, v) only indicates the number of rows and columns where the pixel is located, in order to describe the position of the pixel in the image, an equivalent meaning needs to be established using a physical coordinate system. Thus, a coordinate system parallel to the (u, v) axis is established, with the x-axis corresponding to the u-axis and the y-axis corresponding to the v-axis. (u, v) corresponds to (x, y).

The coordinates are expressed in pixels and correspond to the image's coordinate system, which is measured in millimeters. Thus, the relationship between the position of any pixel in the image and the two coordinate systems can be described as follows:

\begin{equation}\left.\left\{\begin{aligned}&u=\frac{x}{dx}+u_0\\&\nu=\frac{y}{dy}+\nu_0\end{aligned}\right.\right.\end{equation}

The position of the overlap point between the camera optical axis and the image plane in the xy coordinate system is denoted as ($u_0$,$v_0$). By transforming the equation into homogeneous coordinates and matrix form, we can obtain:

\begin{equation}\begin{bmatrix}u\\\nu\\1\end{bmatrix}=\begin{bmatrix}\frac1{dx}&0&u_0\\\\0&\frac1{dy}&\nu_0\\z&0&1\\\end{bmatrix}\end{equation}
where z represents the state vector of the matrix. The inversion is as follows:

\begin{equation}\begin{bmatrix}x\\y\\1\end{bmatrix}=\begin{bmatrix}dx&0&-u_0dx\\0&dy&-\nu_0dy\\0&0&1\end{bmatrix}\begin{bmatrix}u\\v\\1\end{bmatrix}\end{equation}

When a camera captures an object, there exists a unique projection relationship between the object's true coordinates and the obtained image coordinates [8]. Given the knowledge of the camera's geometric parameters, the principle of similar triangles can be used to derive:

\begin{equation}\begin{cases}u=\frac f{-z_e}x_e\\v=\frac f{-z_e}y_e&\end{cases}\end{equation}
Where, $-Z_e$ represents the camera depth, and $f/-Z_e$ represents the scale factor.
Perspective projection only guarantees a one-to-one correspondence between points on the image and points on the projection line. In other words, the same point on the image corresponds to a series of points on the projection line. It is evident that during this process, the depth information of the spatial points is lost, which indicates the need for two or more cameras to determine the three-dimensional object.
By calculating the image coordinates, the position of a three-dimensional point in the camera coordinate system can be obtained:

\begin{equation}\begin{cases}x_e=\frac f{f+w}u\\\\y_e=\frac f{f+w}v\Longrightarrow\\\\z_e=\frac{f^2}{f+w}&\end{cases}\begin{cases}x_e=\frac{-z_e}fu\\\\y_e=\frac{-z_e}fv\\\\z_e=z_e&\end{cases}\end{equation}
where $z_e$ is related to the true distance between the target and the camera. Therefore, this experiment can achieve the conversion from 3D to 2D.

The working scenario designed in this paper is as follows: In a complex work space with great danger, multiple AGVs cooperate with each other to jointly select the best path according to the coordinates of obstacles and target points. In order to improve the likelihood of an AGV successfully executing its task, the robot-generated path needs to meet the following conditions:

(a) After the AGV is started, problems such as insufficient power and sudden failure are ignored;

(b) The paths generated by each AGV need to be collision free from obstacles;

(c) The paths generated between each AGV should be collision-free;

(d) The generated path must ensure that the robot reaches its destination at the same time.

(e) Assume that the speed of each robot i is bounded by $[V_{min}^{i},V_{max}^{i}]$,where $V_{min}^{i}$ and $V_{max}^{i}$ represent the minimum and maximum speed of robot i, respectively.As the robot navigates to the target location, the speed of each robot varies within its velocity boundary.

(f) During the movement of each AGV to the target position, the yaw Angle and travel distance of each AGV need to be kept within its maximum yaw Angle and distance constraints, respectively.

In this experiment, we consider the total path length of the multi-AGV system as the global objective function for the path planning problem. The total path length is defined as the sum of the path lengths of each AGV. Mathematically, the objective function is represented as follows:

\begin{equation}
  F = \min \{ \sum_{i=1}^m L_{i} \}
\end{equation}
where the variable $m$ represents the total number of AGVs, and $L_{i}$ represents the planned path length of the $i$-th AGV. The path length of each AGV, $L_i$, can be calculated as follows:

\begin{equation}
  L_i = \sum_{k=0}^{ns} \mathrm{dis}(p_{i,k},p_{i,k+1})
\end{equation}

We define $p_{i,0}$ and $p_{i,ns+1}$ as the initial and final positions of the $i$-th robot, respectively. The term $\mathrm{dis}(p_{i,k},p_{i,k+1})$ represents the Euclidean distance between waypoints $p_{i,k}$ and $p_{i,k+1}$. The path for the $i$-th robot is denoted as $path_i = [p_{i,0}, p_{i,1}, \dots , p_{i,ns}, p_{i,ns+1}]$, where $path_{i,k}$ ($k=0,1,\dots,n+l$) represents the $k$-th waypoint along the path generated by the $i$-th robot.

The yaw angle, $\theta_{i,k}$, of the $i$-th robot at the $k$-th path segment along the generated path is computed as follows:

\begin{equation}
  \theta_{i,k} = \arccos \left[ \frac{(x_{i,k+1}-x_{i,k})(x_{i,k+2}-x_{i,k+1})+(y_{i,k+1}-y_{i,k})(y_{i,k+2}-y_{i,k+1})}{\mathrm{dis}(p_{i,k},p_{i,k+1})\cdot\mathrm{dis}(p_{i,k+2},p_{i,k+1})} \right]
\end{equation}

The path planning problem is subject to the following constraints:

\begin{equation}
  \begin{aligned}
     & \begin{cases}
         L_i^{\min} \leq L_i \leq L_i^{\max}                                                     \\\
         \theta_i^{\min} \leq \theta_{i,k} \leq \theta_i^{\max}, \quad 1 \leq k \leq ns          \\\
         p_{i,k}p_{i,k+1} \cap \text{obstacle} \in \text{null}, \quad 0 \leq k \leq ns           \\\
         \mathrm{path}_i \cap \mathrm{path}_j \in \text{null}, \quad \forall i \neq j, i,j \in N \\\
         T_i \cap T_j \cap, \dots, \cap T_N \notin \text{non-null}
       \end{cases}
  \end{aligned}
\end{equation}
where $L_i^{\min}$ and $L_i^{\max}$ represent the minimum and maximum path length constraints for the $i$-th robot, respectively. $\theta_i^{\min}$ and $\theta_i^{\max}$ represent the minimum and maximum constraints for the yaw angle rotation of the $i$-th AGV. $T_i$ is the arrival time for the $i$-th robot at the destination position. The variables $x_{i,k}$ and $y_{i,k}$ represent the values of the AGV on the x and y axes, respectively. The term "obstacle" indicates the position coordinate of the obstacle. The term "null" defines the path intersection between any two different robots at the same time as an empty set, ensuring collision-free paths between any AGV and the obstacle. The term "non-null" indicates that the intersection of the arrival times of any two different AGVs is a non-empty set, implying that the paths generated by different AGVs can ensure simultaneous arrival at the destination.

\subsection{PF-DDQN Algorithm}

\begin{figure}[t!]
  \centering
  \includegraphics[scale = 0.5]{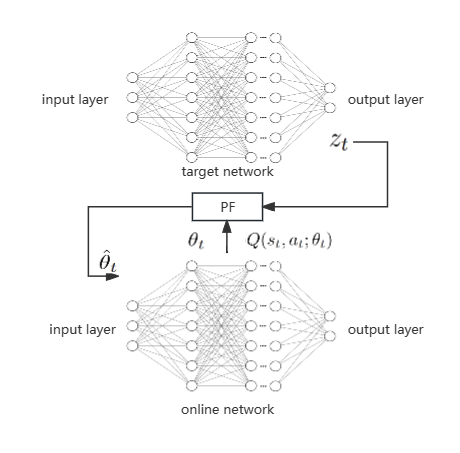}
  \caption{\textbf{DDQN and PF combined structure.}}
  \label{DDQN and PF combined structure.}
\end{figure}

This article proposes the PF-DDQN algorithm, which employs PF to address the issue of weight inaccuracy in deep reinforcement learning models. The algorithm's structure is depicted in Figure \ref{DDQN and PF combined structure.}.

In the proposed algorithm, the state equation and observation equation, which describe the dynamics of the system, are assumed to be represented as follows:

\begin{equation}
  \begin{aligned}
    x_t & = f(x_{t-1}) + w_{t-1} \\
    z_t & = h(x_t) + \nu_t
  \end{aligned}
\end{equation}
where $x_t$ represents the state vector, $z_t$ represents the measurement vector, $f(x_{t-1})$ signifies the state transition function, $h(x_t)$ denotes the transfer function between the state and observation vectors, $w_{t-1}$ represents the process noise, and $v_t$ represents the observation noise.

Subsequently, the weight parameters $\theta_t$, $z_t$, and $Q(s_t,a_t;\theta_t)$ corresponding to time $t$ are substituted into the PF state equation and observation equation, as represented as follows:

\begin{equation}
  \begin{aligned}
    \theta_t & = \theta_{t-1} + W_{t-1}      \\
    Z_t      & = Q(s_t,a_t;\theta_t) + \nu_t
  \end{aligned}
\end{equation}

Let $t-1$ time have a set of posterior particles, expressed as:

\begin{equation}
  \{x_{t-1}(i),\omega_{t-1}(i);i=1,2,\cdots,N\}
\end{equation}
where $N$ represents the number of particles, $x_{t-1}(i)$ denotes the $i$-th particle at time $t-1$, and $\omega_{t-1}(i)$ signifies the weight of the $i$-th particle at time $t-1$.

The entire algorithm flow is described as follows:

\paragraph{Particle set initialization, $t=0$:}

Random samples are drawn from the prior probability density $p(\theta_0)$, denoted as $\theta_0(1)$, $\theta_0(2)$, ..., $\theta_0(N)$ (where $N$ represents the number of random samples).

\paragraph{When $t=1,2,\ldots$, perform the following steps:}

(a) State prediction:

The prior particles at time step $k$ are drawn based on the system's state equation, as shown in the following equation:

\begin{equation}
  \{\theta_{t|t-1}(i);i=1,2,\cdots,N\} \sim p(\theta_{k}|\theta_{k-1})
\end{equation}
(b) Update:

First, the weight update is performed. After obtaining the measured values of the neural network weights, the particle weights, denoted as $w_t^{(j)}$, are calculated based on the system's observation equation as follows:

\begin{equation}
  \omega_{t}^{(i)} = \omega_{t-1}^{(i)}p(Z_{t}|\theta_{t}^{(i)}),\quad i=1,\ldots,N
\end{equation}

Then, calculate the number of effective particles ${{\tilde{N}}_{eff}}$, and compare it with the set threshold $N_{th}$. If ${{\tilde{N}}_{eff}} < N_{th}$, then resample the prior particle set to obtain $N$ particles of equal weight. Otherwise, proceed to the next step.

(c) Estimation:

After iterating $t$ times, the true parameter estimation ${{\hat{\theta }}_{t}}$ is obtained and returned to the estimation network for value function computation. The specific formula is given as follows:

\begin{equation}
  \hat{\theta}_t = \sum_{i=1}^N\theta_{t\mid t-1}(i)\tilde{\omega}_t^{(i)}
\end{equation}

The fundamental PF algorithm maintains the past samples unchanged when sampling at time step $t$, and the importance weights are iteratively computed. In summary, the combination of PF and double deep Q-network involves the following steps:

Step 1: Utilizing the DDQN model, the parameters $\theta_t$,$Q(s_t,a_t;\theta)$, and $z_t$ at time $t$ are employed as inputs to construct the state equation and observation equation for the PF. Furthermore, the particle set is initialized.

Step 2: Iteratively updating, performing state prediction, weight updating, and resampling operations at each time step to obtain the optimal estimation of true weights ${{\hat{\theta }}_{t}}$.

Step 3: Transmitting the optimal true parameters ${{\hat{\theta }}_{t}}$ to the estimation network to obtain $Q(s_t,a_t;\theta)$, which is used to select the action corresponding to the maximum Q-value during experience exploitation, thereby enhancing the accuracy of the neural network's application in the DDQN algorithm. After obtaining $\theta_{t+1}$, $Z_{t+1}$, and $Q(s_{t+1},a_{t+1};\theta)$ at time $t+1$, the above process is repeated to obtain the optimal decision for the next step, cycling through this process until the model converges.

\subsubsection{Status and Reward Mechanisms}

The path planning method for multi-AGV based on PF-DDQN considers each AGV as an intelligent agent, where its position on the grid map is regarded as the controlled object. The actions of the AGV correspond to its movements, and the intelligent agent selects an action based on the current state, executes it, and observes the resulting state and reward. The agent continuously updates its parameters to maximize the reward until the optimal action is determined.

\begin{figure}[t!]
  \centering
  \includegraphics[scale = 0.3]{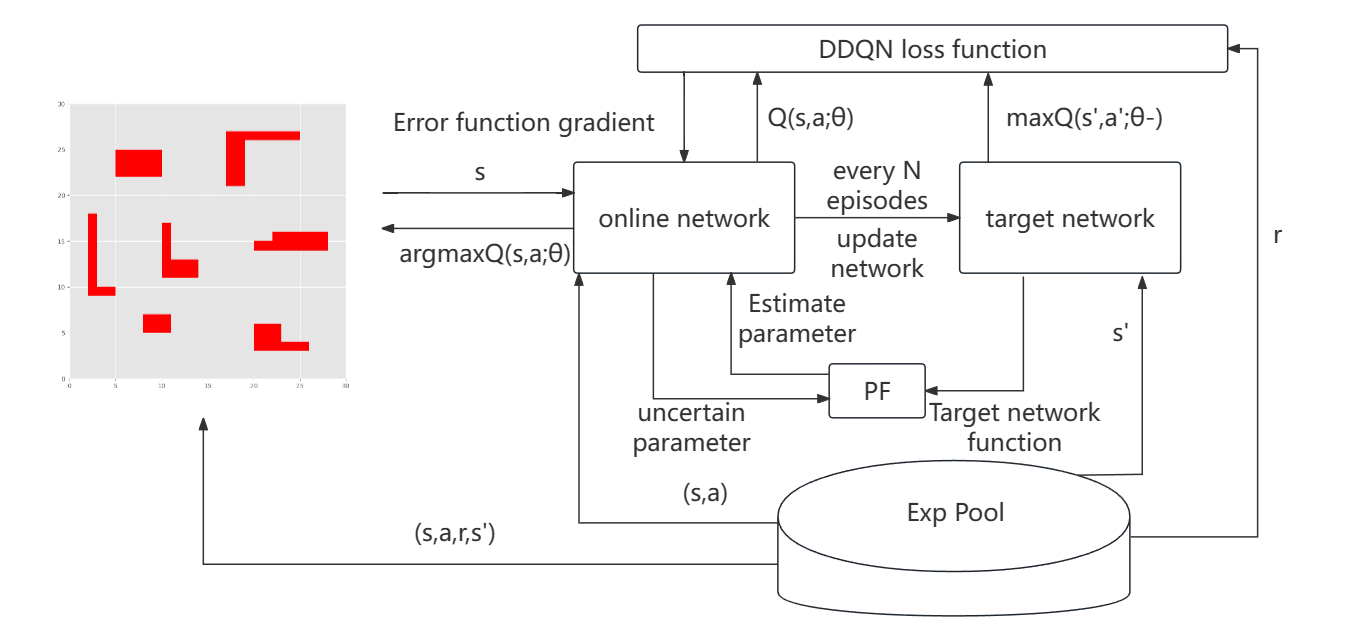}
  \caption{\textbf{Overall framework of the PF-DDQN model.}}
  \label{Overall framework of the PF-DDQN model.}
\end{figure}

The algorithm framework is shown in Figure \ref{Overall framework of the PF-DDQN model.}.The algorithm employs an experience pool to store previous training memories, which aids in the training of the DDQN network. In the framework, $\theta$ represents the network weights at time $t$, $r$ represents the reward at time $t$, $(s, a, r, s')$ represents the replay memory unit at time $t$, $s$ represents the state at time $t$, and $a$ represents the action at time $t$.

\subsubsection{State Space and Action Space}

\begin{figure}[t!]
  \centering
  \includegraphics[scale = 1.5]{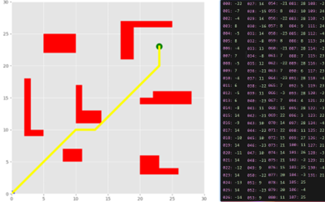}
  \caption{\textbf{Overall planning environment and path diagram.}}
  \label{Overall planning environment and path diagram.}
\end{figure}

In order to accurately represent the operating environment of the AGV and facilitate the training process, this study adopts a grid-based approach to depict the entire working area. Each grid is divided into 2x2m sizes, which corresponds to the length of the AGV. The state variables of the AGV include its relative positions to obstacles and target points. In terms of planning, the AGV has nine possible actions: northward, southward, eastward, westward, northeastward, southeastward, northwestward, southwestward, or remaining stationary. Figure \ref{Overall planning environment and path diagram.} illustrates the overall planning environment and the path diagram.

\subsubsection{Reward Function}

The reward function plays a critical role in reinforcement learning as it guides the learning process of the intelligent agent and influences action selection. Therefore, appropriately defining rewards is essential to achieve desired outcomes and optimal action strategies. In the context of multi-AGV path planning, the primary objective is to minimize the total trajectory points traversed by each AGV while successfully reaching their target points. In this study, the reward is defined with four components:

\paragraph{Baseline Value:} To decrease the overall trajectory count, a penalty value of -4 is assigned when an AGV takes an action to change its position.

\paragraph{Distance to Target Point:} When the sum of distances between all AGVs and their target points decreases, indicating progress, a reward of 5 is assigned. Conversely, a penalty of -5 is given when distances increase.Reward 200 when it overlaps with the target point.

\paragraph{Distance to Obstacle:} To ensure obstacle avoidance, an AGV is penalized with a value of -20 if it collides with an obstacle.

\paragraph{Distance between AGVs:} Rewards are assigned based on the distances between AGVs. Only when AGVs collide, a penalty of -20 is imposed, treating other AGVs as obstacles. This reward and penalty mechanism is independent of the distances between the AGVs.

\subsection{EKF-DDQN Algorithm}

\begin{figure}[t!]
  \centering
  \includegraphics[scale = 0.5]{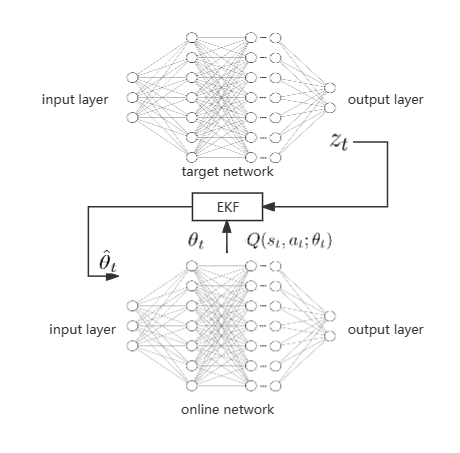}
  \caption{\textbf{DDQN and EKF combined structure.}}
  \label{DDQN and EKF combined structure.}
\end{figure}

\begin{figure}[t!]
  \centering
  \includegraphics[scale=0.2]{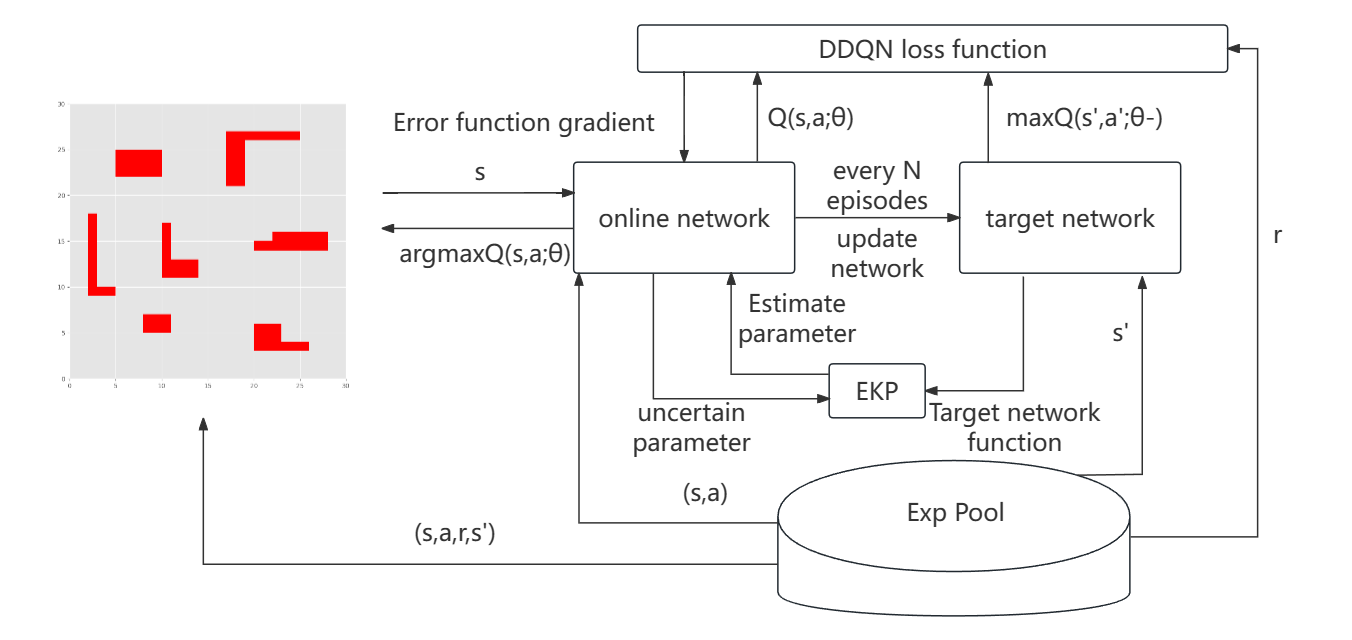}
  \caption{\textbf{Overall framework of the EKF-DDQN model.}}
  \label{Overall framework of the EKF-DDQN model.}
\end{figure}

Similar to PF-DDQN, we can also replace PF with EKF, combine EKF with neural network, and update network weights according to the update mode of EKF. The algorithm structure is shown in Figure \ref{DDQN and EKF combined structure.},The algorithm framework is shown in Figure \ref{Overall framework of the EKF-DDQN model.}.

In the EKF-DDQN algorithm, the state equation and observation equation, which describe the dynamics of the system, are assumed to be represented as follows:

\begin{align}
  x_t & = f(x_{t-1}, u_{t-1}) + w_{t-1} \\
  z_t & = h(x_t) + \nu_t
\end{align}
where $x_t$ represents the state vector, $z_t$ represents the measurement vector, $f(x_{t-1}, u_{t-1})$ signifies the nonlinear state transition function, $h(x_t)$ denotes the measurement function, $w_{t-1}$ represents the process noise, and $\nu_t$ represents the measurement noise.

To update the parameter vector $\theta$ using the EKF, the following EKF update equations are employed:

\begin{align}
  \theta_t^- & = \theta_{t-1}                          \\
  P_t^-      & = F_{t-1}P_{t-1}F_{t-1}^T + Q_{t-1}     \\
  K_t        & = P_t^-H_t^T(H_tP_t^-H_t^T + R_t)^{-1}  \\
  \theta_t   & = \theta_t^- + K_t(z_t - h(\theta_t^-)) \\
  P_t        & = (I - K_tH_t)P_t^-
\end{align}
Where, $\theta_t^-$ and $P_t^-$ represent the predicted parameter estimate and the predicted error covariance matrix, respectively. $F_{t-1}$ and $H_t$ are the Jacobian matrices of the state transition function and the measurement function evaluated at the predicted parameter estimate, while $Q_{t-1}$ and $R_t$ are the process noise covariance matrix and the measurement noise covariance matrix, respectively.

The EKF-DDQN algorithm follows a similar flow to the PF-DDQN algorithm, with the following modifications:

EKF-DDQN Algorithm:

\paragraph{Initialization, $t=0$:}

Set an initial estimate for the parameter vector $\theta_0$ based on prior knowledge or assumptions.

\paragraph{When $t=1,2,...$, perform the following steps:}

(a) State prediction:

Obtain the predicted parameter estimate at time step $t$ using the system's state equation and the EKF prediction equations:

\begin{align}
  \theta_t^- & = f(\theta_{t-1}, u_{t-1})          \\
  P_t^-      & = F_{t-1}P_{t-1}F_{t-1}^T + Q_{t-1}
\end{align}

(b) Update:

Incorporate the new measurement $z_t$ to improve the parameter estimate using the EKF update equations:

\begin{align}
  K_t      & = P_t^-H_t^T(H_tP_t^-H_t^T + R_t)^{-1}  \\
  \theta_t & = \theta_t^- + K_t(z_t - h(\theta_t^-)) \\
  P_t      & = (I - K_tH_t)P_t^-
\end{align}

The remaining steps of the algorithm, including weight updating and resampling, remain unchanged.

The algorithm iteratively performs state prediction and update steps, resulting in an optimized estimate of the true weights $\theta_t$. These optimized weights are then used to compute the value function $Q(s_t,a_t;\theta)$, which is utilized in the decision-making process to select the action corresponding to the maximum Q-value during experience exploitation. The algorithm continues to iterate, updating $\theta_{t+1}$, $z_{t+1}$, and $Q(s_{t+1},a_{t+1};\theta)$ in each iteration.

\section{Experiment Setup and Results}

The numerical simulations in this study were conducted on a system comprising an Intel Core i7-6500 3.20GHz processor, a GPU GTX 3060 with 6GB of memory, and 8GB of RAM. The simulations were performed on both the Ubuntu 20.04 and Windows 11 operating systems using Python.

\begin{figure}[t!]
  \centering
  \includegraphics[scale = 0.3]{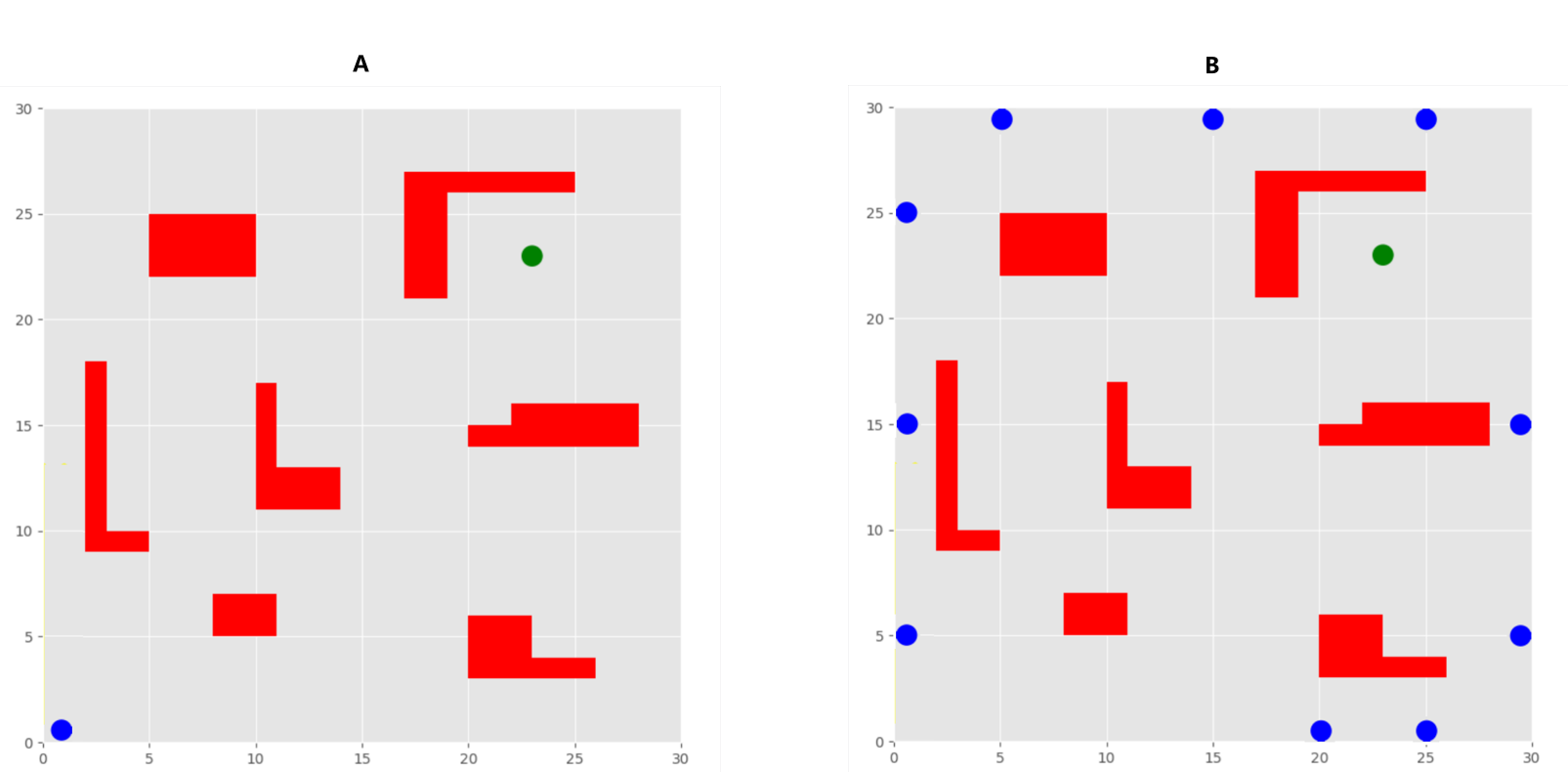}
  \caption{\textbf{Training maps for A) experiments 1 and B) experiment 2,where the robot start point, targets, obstacles, and free space arerepresented as blue, green, magenta,and white.}}
  \label{Training maps for A) experiments 1 and B) experiment 2,where the robot start point, targets, obstacles, and free space arerepresented as blue, green, magenta,and white.}
\end{figure}

This section validates the proposed PF-DDQN path planning algorithm's superior convergence speed and performance compared to the traditional DDQN and EKF-DDQN algorithms through experimental verification. Two simulation experiments are conducted to train PF-DDQN, DDQN, and EKF-DDQN. As illustrated in Figure \ref{Training maps for A) experiments 1 and B) experiment 2,where the robot start point, targets, obstacles, and free space arerepresented as blue, green, magenta,and white.},Experiment 1 focuses on the path planning of a single AGV to a fixed target, where the AGV starts from the initial position and aims to reach a fixed target without colliding with obstacles. Experiment 2 involves the path planning of multiple AGVs to a fixed target, where multiple AGVs start from dispersed initial positions and aim to reach a fixed target without colliding with obstacles or each other. In the same unknown environment and under identical conditions, quantitative and qualitative comparisons are made among PF-DDQN, DDQN, and EKF-DDQN by evaluating their results using the same set of parameters.

In the simulation experiments, the parameters of the deep reinforcement learning network are presented in Table \ref{System Parameters}.

\begin{table}[htbp]
  \centering
  \caption{System Parameters}
  \begin{tabular}{p{4cm}p{1.0cm}}
    \toprule
    Parameter                & Value  \\
    \midrule
    Memory size              & 10000  \\
    Batch size               & 500    \\
    Discount factor          & 0.95   \\
    Learning rate            & 0.0001 \\
    Model update frequency   & 50     \\
    Initial exploration rate & 1.0    \\
    Exploration rate decay   & 0.9995 \\
    Final exploration rate   & 0.001  \\
    \bottomrule
    \label{System Parameters}
  \end{tabular}
\end{table}

\subsection{Experiment 1: Path Planning Comparison for a Single AGV}

In this experiment, the PF-DDQN, EKF-DDQN, and DDQN algorithms underwent 60,000 training iterations. The AGV started from the bottom-left corner of the map and navigated through seven clusters of obstacles to reach a fixed position in the upper-right corner.

Figure 9 depicts the paths planned by the three methods. It is evident from the figure that all three algorithms successfully guide the AGVs to their target points, with similar path lengths. This indicates the feasibility and stability of the algorithms.

The learning curves of the three methods are illustrated in Figure \ref{Learning curves for single-AGV target path planning: Comparison of the proposed method with KF-DDQN and DDQN after training.}. The experimental results demonstrate that the PF-DDQN and EKF-DDQN methods, which incorporate filters, exhibit similar convergence speeds and yield consistent trajectories. In comparison, they outperform the traditional DDQN algorithm by achieving faster and more efficient solutions.

\begin{figure}[t!]
  \centering
  \includegraphics[scale = 1.4]{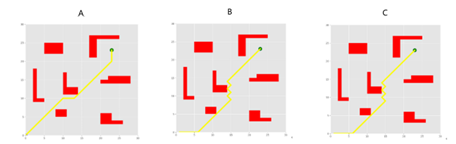}
  \caption{\textbf{Final path results of the experiment for single AGV: A) DDQN, B) EKF-DDQN, and C) PF-DDQN after path smoothing for 60,000 episodes.}}
  \label{Final path results of the experiment for single AGV: A) DDQN, B) EKF-DDQN, and C) PF-DDQN after path smoothing for 60,000 episodes.}
\end{figure}

\begin{figure}[t!]
  \centering
  \includegraphics[scale=0.15]{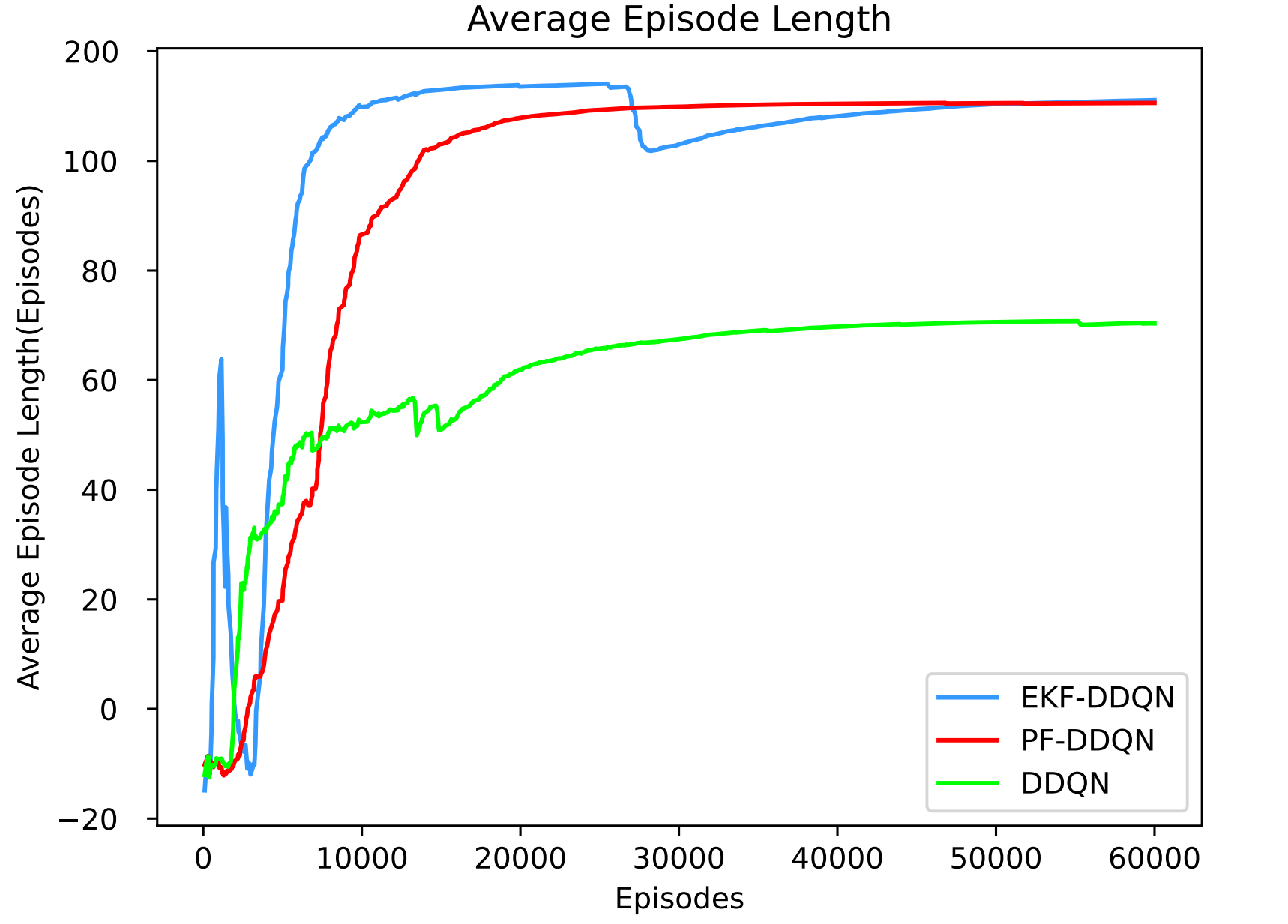}
  \caption{\textbf{Learning curves for single-AGV target path planning: Comparison of the proposed method with EKF-DDQN and DDQN after training.}}
  \label{Learning curves for single-AGV target path planning: Comparison of the proposed method with KF-DDQN and DDQN after training.}
\end{figure}

The DDQN algorithm provides the shortest path from the starting point to the target point, assuming prior knowledge of the map layout. However, in real industrial environments, the behavior of AGVs near obstacles can be hazardous. As the reward function influences path planning behavior, the PF-DDQN method prioritizes safety by better avoiding obstacles, even if it results in a slightly longer path.

Table \ref{Results of Target Path Planning for a Single AGV} presents the specific numerical simulation results of this experiment. Compared to the DDQN algorithm, the proposed method in this study reduced the overall training time by 20.69\% over 60,000 iterations. Additionally, the number of iterations required during the training process decreased by 76.02\% compared to DDQN.

\begin{table}[htbp]
  \centering
  \caption{Results of Target Path Planning for a Single AGV}
  \begin{tabular}{p{2cm}p{3cm} p{3cm}p{3cm}}
    \toprule
                        & DDQN                              & EKF-DDQN                           & PF-DDQN                            \\
    \midrule
    Training Time       & 4.59 Hours                        & 3.95 Hours                         & 3.64 Hours                         \\
    Solution Episode    & After 56,568 episodes/(4.2 Hours) & After 16,482 episodes/(2.79 Hours) & After 13,564 episodes/(2.63 Hours) \\
    Target Hit Times    & 7,498                             & 9,698                              & 25,800                             \\
    Obstacle Hit Times  & 44,598                            & 26,894                             & 4,632                              \\
    Step-Time out Times & 8,579                             & 6,897                              & 4,591                              \\
    Last Path Length    & 23 Grids                          & 30 Grids                           & 30 Grids                           \\
    \bottomrule
    \label{Results of Target Path Planning for a Single AGV}
  \end{tabular}
\end{table}

Since there is only one AGV in this experiment, the constructed network model is relatively simple. Therefore, the PF-DDQN method is not significantly better than the EKF-DDQN method in terms of final reward value. However, compared with PF-DDQN method, the reward curve of EKF-DDQN method has two big changes, which is about to fall into the local optimal solution, indicating that its stability is poor.

The proposed method minimizes unnecessary exploration by the AGV to a great extent by introducing the PF in the process of obtaining target network parameters. This reduces errors resulting from imprecise weight estimates through iterative refinement. Unlike DDQN, this method treats imprecise weights as state variables in the state-space equation, allowing it to converge to the correct range even if the previously learned network parameters are affected by errors. In comparison, DDQN exhibits a broader range of reward values and greater data fluctuations, making convergence less likely.

\subsection{Experiment 2: Comparison of Path Planning for Multiple AGVs}

In this experiment, the PF-DDQN, EKF-DDQN, and DDQN algorithms underwent 60,000 training iterations. Ten AGVs started their journeys from different fixed positions on the map, navigating through seven clusters of obstacles to reach designated fixed positions in the upper-right corner.

Figure \ref{Experiment 2 final path results of in A) DDQN, B) EKF-DDQN and C) PF-DDQN after Multi AGVs path smoothing for 60000 episodes.} illustrates the paths planned by the three methods. It is evident from the figure that the algorithm proposed in this study successfully enables all AGVs to reach their target points, whereas the EKF-DDQN and DDQN algorithms did not achieve the same level of success.

The learning curves of the three methods are depicted in Figure \ref{For multi-AGV target path planning, the proposed method is compared with the learning curve after KF-DDQN and DDQN training.}. The experimental results demonstrate that the improved method exhibits faster convergence compared to the other two methods, with higher values achieved after convergence.

\begin{figure}[t!]
  \centering
  \includegraphics[scale=1.3]{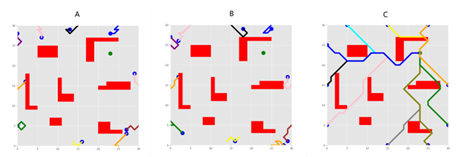}
  \caption{\textbf{Experiment 2 final path results of in A) DDQN, B) EKF-DDQN, and C) PF-DDQN after Multi AGVs path smoothing for 60000 episodes.}}
  \label{Experiment 2 final path results of in A) DDQN, B) EKF-DDQN and C) PF-DDQN after Multi AGVs path smoothing for 60000 episodes.}
\end{figure}

\begin{figure}[t!]
  \centering
  \includegraphics[scale=0.2]{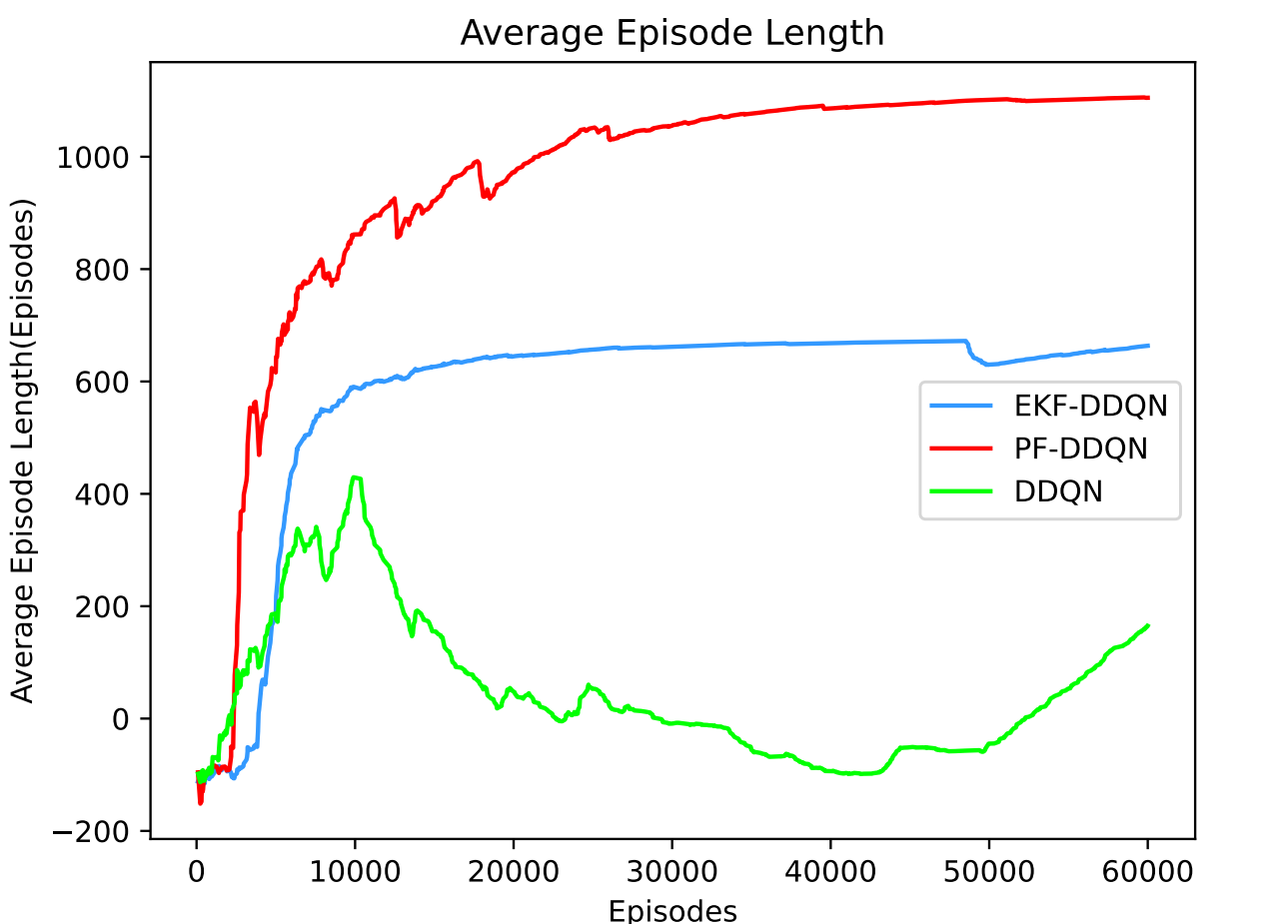}
  \caption{\textbf{For multi-AGV target path planning, the proposed method is compared with the learning curve after EKF-DDQN and DDQN training.}}
  \label{For multi-AGV target path planning, the proposed method is compared with the learning curve after KF-DDQN and DDQN training.}
\end{figure}

Table \ref{Results of target path planning for a single AGV} presents the specific numerical simulation results of this experiment. From the table, it is observed that the DDQN and EKF-DDQN methods enabled the first AGV to reach the target after 16,987 and 27,721 training iterations, respectively. However, when multiple AGVs are involved, the disturbance caused by the neural network's high variance increases exponentially, leading to a rapid decrease in learning efficiency.

As a result, the DDQN method requires a significant amount of time to forget the states associated with individual AGVs receiving high rewards. Conversely, the EKF-DDQN method, due to its highly nonlinear nature, experiences a decrease in accuracy, preventing all AGVs from reaching their target points as intended.

\begin{table}[htbp]
  \centering
  \caption{Results of target path planning for a single AGV}
  \begin{tabular}{p{1.0cm}p{2cm}p{2cm}p{2cm}p{2cm}}
    \toprule
    \multicolumn{5}{c}{DDQN Algorithm (Training Time: 5.14 Hours)}                 \\
    \midrule
    AGV Index & Episode & Training Time & Last Path Length & The Ideal Length \\
    \midrule
    1    & 17989   & 1.3H          & 87               & 24                    \\
    2    & --      & --            & --               & 24                    \\
    3    & --      & --            & --               & 18                    \\
    4    & --      & --            & --               & 14                    \\
    5    & --      & --            & --               & 8                     \\
    6    & --      & --            & --               & 26                    \\
    7    & --      & --            & --               & 24                    \\
    8    & --      & --            & --               & 22                    \\
    9    & --      & --            & --               & 18                    \\
    10   & --      & --            & --               & 9                     \\
    \midrule
    \multicolumn{5}{c}{EKF-DDQN Algorithm (Training Time: 5.74 Hours)}             \\
    \midrule
    AGV Index & Episode & Training Time & Last Path Length & The Ideal Length \\
    \midrule
    1    & 27721   & 3.3H          & 27               & 24                    \\
    2    & --      & --            & --               & 24                    \\
    3    & --      & --            & --               & 18                    \\
    4    & --      & --            & --               & 14                    \\
    5    & --      & --            & --               & 8                     \\
    6    & --      & --            & --               & 26                    \\
    7    & --      & --            & --               & 24                    \\
    8    & --      & --            & --               & 22                    \\
    9    & --      & --            & --               & 18                    \\
    10   & --      & --            & --               & 9                     \\
    \midrule
    \multicolumn{5}{c}{PF-DDQN Algorithm (Training Time: 2.98 Hours)}              \\
    \midrule
    AGV Index & Episode & Training Time & Last Path Length & The Ideal Length \\
    \midrule
    1    & 6584    & 0.96H         & 24               & 24                    \\
    2    & 4783    & 0.67H         & 24               & 24                    \\
    3    & 5869    & 0.72H         & 18               & 18                    \\
    4    & 6202    & 0.78H         & 14               & 14                    \\
    5    & 6563    & 0.81H         & 8                & 8                     \\
    6    & 5905    & 0.73H         & 26               & 26                    \\
    7    & 5968    & 0.74H         & 24               & 24                    \\
    8    & 6330    & 0.79H         & 22               & 22                    \\
    9    & 6097    & 0.76H         & 18               & 18                    \\
    10   & 5780    & 0.71H         & 9                & 9                     \\
    \bottomrule
  \end{tabular}
  \label{Results of target path planning for a single AGV}
\end{table}

In contrast, the PF-DDQN algorithm proposed in this study successfully enables all AGVs to reach their target points. The algorithm converges faster and achieves better results in terms of the path lengths. The training time for the PF-DDQN algorithm is significantly less compared to the other two methods.

In conclusion, the PF-DDQN algorithm outperforms the DDQN and EKF-DDQN algorithms in terms of multi-AGV path planning. It exhibits faster convergence, higher accuracy, and better performance in terms of path lengths.

\section{Conclusion}
To summarize, this study introduces the PF-DDQN method for path planning involving multiple AGVs, which incorporates PF. This approach addresses the limitations of the classical DDQN learning algorithm in noisy and complex environments, while also demonstrating superior fitting accuracy for complex models compared to the use of KF.

The method utilizes a nonlinear model within a neural network to describe the system and integrates PF to estimate the system's state. At each time step, the PF updates the state estimate based on current measurements, while the neural network enhances the accuracy of the estimation. By leveraging the neural network's ability to learn complex patterns in the environment and combining them with the state estimation from PF, the proposed method offers an efficient and effective solution for path planning.

Simulation experiments were conducted to evaluate the performance of the method, revealing significant improvements in training time and path quality compared to the DDQN method, with respective enhancements of 92.62\% and 76.88\%. This research provides valuable insights into path planning and presents a novel and efficient solution applicable to multi-AGV path planning in complex environments. The findings hold promising potential for various applications, including robotics, autonomous vehicles, and unmanned aerial vehicles.

\bibliographystyle{plainnat}
\bibliography{main}

\end{document}